\documentclass[sigconf]{acmart}

\AtBeginDocument{%
  }

\usepackage{graphicx}
\usepackage{array}
\usepackage{multirow}
\usepackage{booktabs}
\usepackage{xcolor}
\usepackage{cleveref}
\usepackage{multicol}
\usepackage{balance}
\usepackage{stfloats} 

\definecolor{gpt2-color}{HTML}{59a88f}
\definecolor{roberta-color}{HTML}{e08379}
\definecolor{t5-color}{HTML}{9f82ba}
\definecolor{llama-color}{HTML}{d67eb4}
\definecolor{mixtral-color}{HTML}{96c44d}

\copyrightyear{2025}
\acmYear{2025}
\setcopyright{cc}
\setcctype{by}
\acmConference[FAccT '25]{The 2025 ACM Conference on Fairness, Accountability, and Transparency}{June 23--26, 2025}{Athens, Greece}
\acmBooktitle{The 2025 ACM Conference on Fairness, Accountability, and Transparency (FAccT '25), June 23--26, 2025, Athens, Greece}\acmDOI{10.1145/3715275.3732112}
\acmISBN{979-8-4007-1482-5/2025/06}

\begin{document}

\title[Gender Trouble in Language Models]{Gender Trouble in Language Models: An Empirical Audit Guided by Gender Performativity Theory}

\author{Franziska Sofia Hafner}
\affiliation{%
  \institution{University of Oxford}
  \country{} 
  \city{}
  }
\email{Franziska.Hafner@oii.ox.ac.uk}

\author{Ana Valdivia}
\affiliation{%
 \institution{University of Oxford}
 \country{}
 \city{}
 }
\affiliation{%
\institution{Institute of Advanced Studies (UCL)}
\country{}
\city{}
}
 \email{Ana.Valdivia@oii.ox.ac.uk}

\author{Luc Rocher}
\affiliation{%
  \institution{University of Oxford}
  \country{} 
  \city{}
}
\email{Luc.Rocher@oii.ox.ac.uk}

\renewcommand{\shortauthors}{Hafner et al.}

\begin{abstract}
Language models encode and subsequently perpetuate harmful gendered stereotypes.
Research has succeeded in mitigating some of these harms, e.g. by dissociating non-gendered terms such as occupations from gendered terms such as `woman' and `man'. This approach, however, remains superficial given that associations are only one form of prejudice through which gendered harms arise. Critical scholarship on gender, such as gender performativity theory, emphasizes how harms often arise from the construction of gender itself, such as conflating gender with biological sex. In language models, these issues could lead to the erasure of transgender and gender diverse identities and cause harms in downstream applications, from misgendering users to misdiagnosing patients based on wrong assumptions about their anatomy.

For FAccT research on gendered harms to go beyond superficial linguistic associations, we advocate for a broader definition of ‘gender bias’ in language models. We operationalize insights on the construction of gender through language from gender studies literature
and then empirically test how 16 language models of different architectures, training datasets, and model sizes encode gender. We find that language models tend to encode gender as a binary category tied to biological sex, and that gendered terms that do not neatly fall into one of these binary categories are erased and pathologized. Finally, we show that larger models, which achieve better results on performance benchmarks, learn stronger associations between gender and sex, further reinforcing a narrow understanding of gender. Our findings lead us to call for a re-evaluation of how gendered harms in language models are defined and addressed.
\\

\noindent\textit{Warning: This article discusses the erasure and pathologization of transgender and gender diverse identities in academic literature and in language models.}

\end{abstract}



\maketitle

\section{Introduction}

Language plays a key role in constructing, reinforcing, and subverting gender norms~\citep{baker2008}. 
The English language, for example, constructs gender by assigning the binary categories `boy' and `girl' to babies at birth, reinforces gender through phrases such as `boys will be boys' and `throw like a girl', and subverts gender through terms not fitting into the `boy/girl' binary, such as the use of `they/them' as singular pronouns. 
The more frequently some gender norm is expressed through language, the more it is normalized~\citep{butler_gender_2006}. Conversely, gendered experiences not often expressed through language might be perceived as abnormal, unacceptable, or even unimaginable.
Language models (LMs) encode these norms from the books, online news articles, and online forums that make up their training data~\citep{birhane2024, dodge2021,steed2022,feng2023,koksal2023}. Encoding human biases and their representation in popular culture, LM-based technologies can then persistently reproduce prejudice against sexual orientation or gender identity.

In an effort to mitigate bias and discrimination in LMs, computer scientists have audited LMs for harmful stereotypes. This research has contributed to understanding and reducing these harms~\citep{gallegos2024}.
For example, researchers have
used large dictionaries of ‘bias probes’ and ‘prefix templates’ to elicit biased sentence completions~\citep{solaiman2019,sheng2019}. Further research on downstream applications, notably in HR and recruitment as well as in clinical and medical applications, showed that LMs consistently stereotype certain races, ethnicities, and genders.
Subsequently, new approaches have been developed which target training datasets (e.g. by re-balancing data with pre-defined list of biased word pairs such as ‘he/she’ and ‘king/queen’~\citep{liu2021}), learning phase (e.g. by adding an equalizing regularization term to the loss function that produces embeddings with less association between ‘he/she’ pronouns and occupations and by using reinforcement learning with human feedback), or post-processing phase (e.g. by setting prompts that help avoid biased sentence completion, or replacing stereotypical keywords)~\citep{gallegos2024}. 

However, while debiasing efforts reduce easy-to-see forms of bias in simple benchmarks, they fail to address more complex and deeply ingrained associations~\citep{wang2023,ovalle2024}, especially in adversarial settings~\citep{qi2023} or in Chain of Thought settings that prompt models to ‘think’~\citep{shaikh2023}, both circumventing debiasing efforts. \citet{hofmann2024} found that such efforts can successfully reduce negative overt racist stereotypes but exacerbate covert stereotypes, in the form of dialect prejudice.
For example, some models are optimized to reduce the association of ‘he/she’ pronouns or ‘black/white’ racial categories with occupations. This may not affect cases where gender is inputed from pronouns other than ‘he/she’, or race from racialised terms other than ‘black/white’, e.g. through implicit information such as dialect or writing style.
Moreover, models optimized according to specific categories such as `man' and `woman' risk essentializing complex social identities such as gender into a narrow set of biological attributes or culturally dominant norms, by reinforcing the idea that these categories are fixed, universal, or exhaustive.

Addressing bias and discrimination in LMs requires moving beyond aligning models with pre-existing human preferences, by leveraging decades of critical research from the social sciences on how social characteristics are and should be represented in language~\citep{blodgett2020}.
Research in fields such as gender studies, sociolinguistics, feminist studies, and queer studies highlights the importance of language not merely in representing gender, but also in constructing, reinforcing, and subverting gender norms~\citep{baker2008}. 
Critically, they show that gendered harms often arise from the construction of gender itself, instead of from stereotypical linguistic associations. For example, gender and sex are frequently linked narrowly in language (\textit{man} linked to \textit{male}, \textit{woman} linked to \textit{female}). This strong association can be harmful both to cisgender people (e.g., if the ability to conceive is narrowly tied to womanhood, this is harmful to cisgender women wo do not want to or cannot get pregnant) and to transgender and gender diverse people (e.g., a transgender woman who cannot get pregnant might be excluded from womanhood on this basis).
Additionally, by defining classifications of gender (e.g., \textit{man} and \textit{woman}), language can contribute to the erasure of gender identities that are not considered in such classifications, such as transgender and gender diverse people. We build on these insights from gender studies, and apply them to the study of LMs.

We derive three main insights on the importance of language from Butler's works on gender performativity and apply them to LMs:
(a) \textit{distinction between gender and sex} (LMs should not essentialize gender by conflating it with biological sex),
(b) \textit{meaningful embeddings of diverse genders} (LMs should provide meaningful embeddings for all gender identity terms), and
(c) \textit{non-stereotypical representations} (LMs should not encode harmful stereotypes about any gender identity terms).

We test how 16 LMs, including versions of Llama, Mistral, RoBERTa, T5, and GPT, construct gender, by measuring the conditional probability they assign in model predictions across different contexts related to sex, gender, and illnesses.
Firstly, we find that most of the tested models appear to follow essentializing patterns regarding the relationship between gendered words and sex characteristics (e.g., associating `man' with testosterone and `woman' with estrogen). Crucially, we find that gender and sex are conflated more as models get larger.
Secondly, some of the models consistently associate unrelated non-human words with sex characteristics more strongly than they do words associated with transgender and gender diverse people, such as the words `nonbinary', `transgender', `genderqueer', `genderfluid', and `two-spirit'. This indicates a lack of meaningful embeddings for these words.
Finally, most models appear to encode harmful pathologizing associations for transgender and gender diverse words, in particular by associating them with mental illness.

Left unchecked, the essentializing and pathologizing issues we uncovered could have harmful consequences in numerous settings.
Our findings affirm the need to consider concepts such as gender beyond simple alignment practices, by engaging critically with them throughout the entire model development process. Ultimately, we hope that efforts to investigate how various social attributes are conceptualized by LMs can lead to the development of models that do not perpetuate, but mitigate, the harms associated with narrow and essentialist definitions.

\section{Background}

\subsection{The Construction of Sex and Gender through Language}\label{background_gender_sex}

Gender and sex are frequently not recognized as separate concepts in discourse on gender. Instead, gender is often framed as binary (every person fits into one of the distinct categories `man' or `woman'), immutable (this category can not change), and physiological (this category is assigned based on physical characteristics)~\citep[p. 88]{keyes_misgendering_2018}. This binary, immutable, and physiological understanding of gender is called the `folk understanding' ~\citep{keyes_misgendering_2018}. While this understanding of gender is not universal across cultures or historical periods, it is dominant in many Western and English-speaking contexts.

The folk understanding of gender has been extensively critiqued in gender studies. In particular, scholars have highlighted how sex and gender are not inherently linked, but become interlinked through their social construction~\citep{wittig1985mark}. 
As Simone de Beauvoir writes in \textit{The Second Sex} on the construction of gender, `one is not born, but rather becomes, a woman'~\citep[p.~267]{debeauvoir1953second}. 
Similarly, Judith Butler 
challenges the innateness of sex by asking `is it natural, anatomical, chromosomal, or hormonal'~\citep[p.~9]{butler_gender_2006}. 
This untangling of the elements that make up a supposedly binary sex leads Butler to question:
`if the immutable character of sex is contested, perhaps this construct called ``sex'' is as culturally constructed as gender'~\citep[p.~9]{butler_gender_2006}.
Instead of portraying sex as a precursor to gender (as does the folk understanding), Butler therefore argues that 
gender and sex become interlinked in their social construction~\citep{butler_bodies_2011}.

According to gender theorists, language plays a crucial role in the interlinked construction of sex and gender. 
By describing concepts, language allows us to imagine them more easily. Therefore, the power of language lies in creating an `imaginable domain'~\citep[p.~12]{butler_gender_2006}. 
By providing ways to describe some gendered experiences but not others, language can limit which genders and gender expressions are considered normal, acceptable, and even possible. 
For example, the English language provides many words for articulating binary sex and gender categories (\textit{the male and the man, the female and the woman}). Through this, `language gains the power to create ``the socially real'''~\citep[p.~156]{butler_gender_2006}. 

\subsection{The Pathologization of Sex and Gender outside a Binary}\label{background_pathologisation}

When individuals' bodies and gender expressions do not align with the sexed and gendered norms defined through language, individuals face significant social sanctions. Foucault, for example, illustrates this in the book \textit{Herculine Barbin}, where he retraced the struggles of a French intersex person born in 1838. 
As their body did not fit into a `male' nor `female' norm, legal and medical institutions of the time were not able to categorize Barbin within the gender and sex binary~\citep{foucault1980herculine}. 
This led to a series of institutional interventions, including medical examinations and legal debates over how to determine someones `true' sex, hidden behind `anatomical deceptions'~\citep[p. 8]{foucault1980herculine}. To this day, intersex babies are still routinely operated to fit into a `male' or `female' category~\citep{morgenroth2018gender}.

In the 20th century, frameworks such as the Diagnostic and Statistical Manual of Mental Disorders (DSM) and International Classification of Diseases (ICD) classified gender variance as `disordered'~\citep{drescher_minding_2012, drescher_queer_2015}. These categorizations both entrenched stigma and positioned medical institutions as gatekeepers of legitimacy and care~\citep{inch_changing_2016, davy_politics_2015, dewey_dys_2017}. While terminology evolved from the diagnosis of a `gender identity disorder' to `gender dysphoria' in the 2013 DSM edition (DSM-5), scholars nevertheless argue that the structural underpinnings of gatekeeping persist \citep{inch_changing_2016, davy_democratising_2018}.

Narrow norms for acceptable expressions of sex and gender are enforced by pathologizing sex and gender outside a binary.
According to Ian Hacking, the 
historical and 
ongoing role of medical institutions in defining and enforcing gender norms lies, at least partially, in its ability to construct scientific classifications.
These classifications `may bring into being a new kind of person'~\citep[p.~285]{hacking2007kinds}. Scientific knowledge, rooted in social, medical, and biological Western sciences, has historically created and reinforced the folk understanding of gender in its scientific classifications, by rigidly linking gender to sex. As a result, Hacking suggests that these classifications create stereotypes which, in the case of the folk understanding of gender, lead to the pathologization of non-normative gender expressions and reinforce an exclusionary imaginable domain of gender identity.

\subsection{Folk Understanding of Gender in Previous Evaluations of Language Models}

A decade ago, the popular experiment \textit{king}+\textit{woman}--\textit{man} $\approx$ \textit{queen} illustrated the potential of Natural Language Processing (NLP) in capturing lexical relations through LMs~\citep{vylomova2015take}. Concretely, the experiment demonstrated that the vectorial representation of the word \textit{king}, when summed with \textit{woman} and subtracted by \textit{man}, resulted in the representation of \textit{queen}\footnote{This example is not meant as an illustration of bias, but rather as an illustration of how word embeddings can model semantic relationships.}.

Researchers have since used the vectorial representation of words to uncover biased and stereotypical representations of gender in LMs. 
\citeauthor{bolukbasi_man_2016} have notably discovered that, similar to the association between `woman' and `queen', there also exist other gendered associations in word embeddings, such as  \textit{computer programmer}+\textit{woman}--\textit{man} $\approx$ \textit{homemaker}. \citeauthor{bolukbasi_man_2016} consider the association between \textit{woman} and \textit{homemaker} to be inappropriate, and propose a method for removing such associations from word embeddings. Conversely, they find the association \textit{prostate cancer}+\textit{she}--\textit{he} $\approx$ \textit{ovarian cancer} to be appropriate, and explicitly ensure this remains in the embeddings~\citep[p.~2]{bolukbasi_man_2016}. 
This choice, however, conflates biological sex characteristics with linguistic gender, implying that someone with ovaries cannot be associated with the pronoun \textit{he}, or someone with a prostate cannot be associated with \textit{she}. Such assumptions reinforce a folk understanding of gender by reducing gender to a binary construct (\textit{he}/\textit{she}) and aligning these categories with biological sex characteristics.

Despite an extensive focus on removing gender bias from language models, problematic assumptions such as the folk understanding of gender are rarely questioned.
For example, \citet{devinney_theories_2022} surveyed 176 articles that aim to identify or mitigate gender bias in NLP systems. Over 93\% of the articles they surveyed operationalized gender as a binary category. While some mention this as a limitation of their research, only 7\% of the surveyed articles provide a proper definition of gender that is inclusive of transgender and/or nonbinary individuals, with more than half of these articles not extending their analysis to explicitly reflect this.

\subsection{Previous Critical Evaluations of Gender in Language Models}

The relatively few articles on gender bias in NLP that operationalize gender beyond a folk understanding are concerned primarily with addressing anti-LGBTQIA+ bias~\citep{felkner_winoqueer_2023, ovalle2023m, queerinai2023queer, dhingra2023queer}. While critical gender theory and queer studies overlap greatly in the social sciences, there exists a notable divide in FAccT and NLP literature. Most gender bias literature addresses performance differences between binary cisgender women and men, while research addressing bias against non-cisgender individuals is labeled as LGBTQIA+-related and discussed separately. However, this separation ignores the great intersection between both gender and anti-LGBTQIA+ bias, which both stem from similar patriarchal structures~\citep{zhou_queer_2024}. Additionally, this separation fails to recognize that the folk understanding of gender is not just an issue within the LGBTQIA+ community, but can also harm cisgender heterosexual individuals. For example, the ways in which the folk understanding of gender equates the ability to conceive with womanhood is not just harmful to transgender people, but also to cisgender women who, for various reasons, are not able to or do not want to get pregnant. Thus, we believe that a more nuanced understanding of gender is important for all work on gender bias in NLP, not just such work that focuses specifically on the LGBTQIA+ community.\\


This article contributes to the FAccT and NLP literature by introducing an empirical framework for testing how LMs encode and reproduce the folk understanding of gender.
While previous work has focused on how gendered stereotypes are reproduced by LMs, we go further by illustrating how gender itself is constructed in LMs. This aligns with Butler's theories on the interplay of gender, language and power. Just as gender theorists argue that language not only reflects, but also actively constructs, gender by associating it narrowly with sex, LMs might also contribute to the continued reproduction of narrow and exclusionary concepts of gender through encoding this folk understanding. 
Moreover, our empirical results reveal that this perpetuation leads to the pathologization of gender in LMs, with clear patterns linking gendered terms that subvert the folk understanding to mental illness. This underscores the urgent need for interdisciplinary scholarship at the crossroads of NLP and gender studies. Such collaboration can provide a deeper critique of how computational models continue to reinforce cultural norms and assumptions regarding gender.

\section{Methods}

\subsection{Measuring Conditional Probabilities from Language Models}\label{prompting_methods}

We probe implicit associations between gendered and sexed words, as well as associations between gendered words and physical or mental illnesses. To investigate these associations made by LMs, we use a standard approach by comparing, for each tested model, the probability distribution assigned to short stereotypical sentences~\cite{gallegos2024}.

To test both autoregressive and masked LMs, we design these sentences in a standardized manner, starting with a context (e.g. `The person who has testosterone is') followed by word(s) of interest (e.g. `a woman'). Autoregressive models such as GPT2 are trained to predict the next token after a sequence of tokens, while masked models such as RoBERTa are trained to predict masked tokens from preceding and subsequent tokens in a sequence. Crafting all sentences with a context followed by a word of interest allows us to calculate, for each model and each sentence, the probability of the predicted word of interest conditional on the context. If the predicted word of interest is split over multiple tokens, we calculate the joint probability over all tokens (using the sum of log probabilities)~\citep{hofmann2024}. For example, if the word `nonbinary' is tokenized as `non-bi-nary', we sum the log probabilities of these three tokens. 

Using this general approach, we construct two sets of sentences. We designed the first set to investigate how gender is associated with sex in LMs. This mirrors how social theorists argue that gender is frequently constructed through sex in natural language (see~\Cref{background_gender_sex}). Thus, we are interested in the probability of different gendered predictions, conditioned on the sex-related context of a sentence. We designed the second set of sentences to investigate how illness is associated with genders that subvert norms. This mirrors how social theorists argue that non-normative expressions of gender often lead to pathologization in society (see~\Cref{background_pathologisation}). Thus, we are interested in the probability of different illness-related predictions, conditioned on the gender-related context of a sentence.

\subsubsection{Sex-Gender Association and Non-Human Baseline}
\label{subsub:sex-gender-baseline}

We define the conditional probability \( \mathbb{P}(g \mid \textsf{Context}(s); \theta) \) as the probability that a model $\theta$ outputs gender identifier \( g \) after the context:
\[ \textsf{Context}(s) \coloneq \text{"The person who [is/has/has a] } s \text{ is"}\]
where \( s \) is a sex characteristic from the set \( S = \) \{male, penis, prostate, testosterone, XY chromosomes, female, vagina, uterus, estrogen, XX chromosomes\}, %
and \( g \) is a gender identifier from the set \( G = \) \{a man, a woman, transgender, nonbinary, genderqueer, genderfluid, two-spirit\}.
For instance, \( \mathbb{P}(\text{a man} \mid \textsf{Context}(\text{testosterone}); \theta) \) is the probability that a model $\theta$ assigns to ``a man'' in the sentence “The person who has testosterone is a man”.

In order to determine the degree to which each model learns to associate sex characteristics with gender identifiers, we also measured conditional probabilities on unrelated, non-human words of interest.
We generated a list of 50 random nouns between 9 and 13 characters using the Python library wonderwords.\footnote{\url{https://github.com/mrmaxguns/wonderwordsmodule}} We excluded three human-related words (bartender, instructor, and creationist), resulting in 47 non-human nouns. The full list of non-human baseline words we used can be found in Appendix~\ref{appendix_probing_terms}. Using the same approach as for gender-sex associations, we then measured the conditional probability of each non-human noun following each sex characteristics context \(\textsf{Context}(s)\).

\subsubsection{Gender-Illness Association}

We define the conditional probability \( \mathbb{P}(i \mid \textsf{Context}(g); \theta) \) as the probability that a model $\theta$ outputs illness \( i \) after the context:
\[ \textsf{Context}(g) \coloneq \text{"The person who is} \ g \ \text{has}\text{"}\]
where \( g \) is a gender identifier from the set \( G = \) \{a man, a woman, transgender, nonbinary, genderqueer, genderfluid, two-spirit\}, and \( i \) an illness from a comprehensive set of 110 illness-related words derived from physical illnesses listed in the Global Burden of Disease Study 2021 and from mental illnesses recognized by the American Psychological Association (for a complete list of illness terms, see Appendix~\ref{appendix_probing_terms}).
For instance, \( \mathbb{P}(\text{anxiety} \mid \textsf{Context}(\text{transgender}); \theta) \) is the probability that a model $\theta$ assigns to `anxiety' in the sentence `The person who is transgender has anxiety'.

\subsection{Comparing Probabilities using Log Probability Ratios}

To compare how the probabilities of gendered and illness-related completions change depending on sexed and gendered contexts, we use log probability ratios (LPR). LPRs are a common metric used to compare probabilities extracted from LMs. They are particularly useful in highlighting the degree of change in very small probabilities. In general, a log probability ratio is expressed as \( \text{LPR} =\log \mathbb{P}(\text{A})  - \log \mathbb{P}(\text{B}) \). The ratio is positive if the probability of A is greater than the probability of B.

We calculate three different LPRs. The first compares the probability between different gendered predictions given a specific sexed context. 
This allows us to investigate how the probabilities of predictions that align with common associations between sex and gender change as models get larger.
The second compares the probability of a specific gendered prediction given differently sexed contexts. 
This allows us to investigate how specific sex characteristics lead to differently gendered predictions.
The third compares the probability of a specific illness-related prediction, given differently gendered contexts. 
This allows us to investigate how specific gendered contexts lead to different pathologization in predictions. 

\subsubsection{Folk-Subversive Log Probability Ratio} \label{methods_folk_subversive}
To test whether a model $\theta$ learns folk associations between gender and sex, we compare the probabilities of folk associations of gender and sex (e.g. `The person who has testosterone is a man') with the probabilities of subversive associations (e.g. `The person who has testosterone is nonbinary'). The Folk--Subversive log probability ratio can be expressed as:
\begin{equation}\label{eq:Folk-Subversive}
\begin{aligned}
&\text{Folk--Subversive LPR}(\theta) \coloneq \\ 
&\quad \frac{1}{|G|} \sum_{g \in G} \sum_{s \in S} \delta_{\textsf{FOLK}}(g, s) \cdot \log\left( \mathbb{P}(g \mid \textsf{Context}(s); \theta) \right)
\end{aligned}
\end{equation}
where \( \delta_{\textsf{FOLK}}(g, s) \) indicates if $g$ and $s$ align with the folk understanding (\(+1\) for `a man' in the five contexts with male sex characteristics, and or `a woman' in the five contexts with female sex characteristics) or do not align (\(-1\)  for `a woman', `nonbinary', `transgender', `genderqueer', `genderfluid', and `two-spirit' in contexts with male sex characteristics, or `a man', `nonbinary', `transgender', `genderqueer', `genderfluid', and `two-spirit' in contexts with female sex characteristics). A value greater than zero indicates folk understanding of gender while a value lesser than zero indicates subversive understanding of gender.

\begin{table*}[b]
    \centering
    \begin{tabular}{ >{\raggedright}m{2cm}  >{\raggedright\arraybackslash}m{2.5cm}  >{\raggedright}m{5cm}  >{\raggedright}m{3cm}  >{\raggedleft\arraybackslash}m{1cm} }
        \hline
        \textbf{Model Category} & \textbf{Transformer Architecture} & \textbf{Training Dataset} & \textbf{Specific Model} & \textbf{Model Size} \\
        \hline
        \multirow{2}{*}{\textcolor{roberta-color}{RoBERTa} \cite{liu2019RoBERTarobustlyoptimizedbert}} & \multirow{2}{3.5cm}{Encoder-only} & \multirow{2}{5cm}{BookCorpus, Wikipedia, CC-News, OpenWebText, Stories} & RoBERTa-base & 125M \\
        \cline{4-5}
         &  &  & RoBERTa-large & 355M \\
        \hline
        \multirow{4}{*}{\textcolor{gpt2-color}{GPT2} \cite{radford_language_2019}} & \multirow{4}{3.5cm}{Decoder-only\\ Autoregressive} & \multirow{4}{5cm}{WebText} & GPT-2 & 124M \\
        \cline{4-5}
         &  &  & GPT-2 Medium & 355M \\
        \cline{4-5}
         &  &  & GPT-2 Large & 774M \\
        \cline{4-5}
         &  &  & GPT-2 XL & 1.5B \\
        \hline
        \multirow{4}{*}{\textcolor{t5-color}{T5} \cite{raffel2023exploringlimitstransferlearning}} & \multirow{4}{3.5cm}{Encoder-Decoder} & \multirow{4}{5cm}{Colossal Clean Crawled Corpus (C4)} & T5 Small & 60.5M \\
        \cline{4-5}
         &  &  & T5 Base & 223M \\
        \cline{4-5}
         &  &  & T5 Large & 738M \\
        \cline{4-5}
         &  &  & T5 3B & 2.85B \\
        \hline
        
        \multirow{2}{*}{\textcolor{mixtral-color}{Mistral} \cite{jiang2023mistral7b}} & \multirow{2}{3.5cm}{Decoder-only\\ Autoregressive} & \multirow{2}{3cm}{Undisclosed} & Mistral-7B-v0.3 & 7.25B \\
        \cline{4-5}
         &  &  & Mixtral-8x7B-v0.1 & 46.7B \\
        \hline
        \multirow{4}{*}{\textcolor{llama-color}{Llama} \cite{touvron2023llamaopenefficientfoundation}} & \multirow{4}{3.5cm}{Decoder-only\\ Autoregressive} & \multirow{4}{5cm}{English CommonCrawl, C4, Github, Wikipedia, Gutenberg and Books3, ArXiv, Stack Exchange} & Llama-3.2-1B & 1.24B \\
        \cline{4-5}
         &  &  & Llama-3.2-3B & 3.21B \\
        \cline{4-5}
         &  &  & Llama-3.1-8B & 8.03B \\
        \cline{4-5}
         &  &  & Llama-3.1-70B & 70.6B \\
        \hline
    \end{tabular}
    \caption{\textbf{Overview of the Evaluated Language Models}}
    \label{tab:models}
\end{table*}

\subsubsection{Sex--Gender Log Probability Ratio}
\label{subsub:sex-gender-lpr}

To test how sexed contexts influence gendered predictions, we compare the probability of a specific gendered prediction (e.g., `a man'), when conditioned on two differently sexed contexts (e.g., `The person who has testosterone is' vs. `The person who has estrogen is'). 
This approach involves using matched contexts that align male and female sex characteristics, then comparing the probability of specific predictions.
Termed `matched guise probing', this technique of matching contexts and subsequently comparing the probability of a specific prediction was proposed by \citet{hofmann2024}. It allows us to analyze how gendered associations vary based on the sexed information within a prompt. This method reflects the societal process of assigning gender based on (perceived) sex characteristics.

We compare the conditional probability of a gendered prediction given two `matched' sexed contexts using a LPR expressed as:
\begin{equation}\label{eq:Sex-Gender}
\begin{aligned}
\text{Sex--Gender LPR}(s, g; \theta) \coloneq\ 
&\log\left( \mathbb{P}(g \mid \textsf{Context}(s); \theta) \right) \\
&\quad - \log\left( \mathbb{P}(g \mid \textsf{Context}(s'); \theta) \right)
\end{aligned}
\end{equation}
where \(s \in S_F\), the set of female characteristics (female, vagina, uterus, estrogen, XX chromosomes), and \(s'\) its matched characteristics in \(S_M\), the set of male characteristics (male, penis, prostate, testosterone, XY chromosomes).
\(\text{Sex--Gender LPR}(s, g; \theta)\) is positive if the probability of gendered prediction $g$ is greater in the female context $s$ compared to the matched male context $s'$---and negative otherwise.

\subsubsection{Gender--Illness Log Probability Ratio}
To test how gendered contexts influence illness-related predictions, we compare the probability of a specific mental or physical illness-related prediction (e.g., `depression'), when conditioned on two differently gendered contexts (e.g., `The person who is a man has' vs. `The person who is nonbinary has'). We use the same ‘matched guised probing’ approach as for Sex--Gender LPR (see \Cref{subsub:sex-gender-lpr}) with two matching contexts. We match sentences containing the gendered contexts `a woman', `nonbinary', `transgender', `genderqueer', `genderfluid', and `two-spirit' to an otherwise equivalent sentence with the context `a man'. This allows us to analyze how illness-related associations vary based on gendered information within a context. This method reflects the societal process of pathologizing individuals based on their (perceived) gender or gender-nonconformity.

We compare the conditional probability of an illness-related prediction given two `matched' gendered contexts using a LPR, Gender--Illness LPR, expressed as:
\begin{equation}\label{eq:Gender-Illness}
\begin{aligned}
\text{Gender--Illness LPR}(g, i; \theta) \coloneq\ 
&\log\left( \mathbb{P}(i \mid \textsf{Context}(g); \theta) \right) \\
&\quad - \log\left( \mathbb{P}(i \mid \textsf{Context}(g'); \theta) \right)
\end{aligned}
\end{equation}
where $g$ is a gender identifier from the set \{a woman, transgender, nonbinary, genderqueer, genderfluid, two-spirit\} and \(g' = \text{`a man'}\) (the matched gender). We test every illness \( i \) from a comprehensive set of 110 illness-related words derived from physical illnesses listed in the Global Burden of Disease Study 2021 and mental illnesses recognized by the American Psychological Association (see Appendix~\ref{appendix_probing_terms}). \(\text{Gender--Illness LPR}(g, i; \theta)\) is positive if the probability of the illness-related prediction \(i\) is greater in the gendered context \(g\) than in the context mentioning `a man'---and negative otherwise.

\section{Results}

We empirically test how LMs conceptualize gender using sixteen pre-trained models, representing different architectures, training datasets, and parameter sizes. To ensure reproducibility and replicability, we only included publicly available open source models (four GPT2 models, two RoBERTa models, four T5 models, four Llama models, and two Mistral models; see \Cref{tab:models}). Model sizes range from 60.5 million parameters for the smallest T5 model to 70.6 billion parameters for the largest Llama model, allowing us to investigate how constructions of gender are learned by models as they get larger. For each model, we tested both asociations between gender and sex as well as pathologization of trans and gender diverse identities.

\subsection{Associations between Gender and Sex}

\begin{figure*}[b]
\centering
\includegraphics[width=\linewidth]{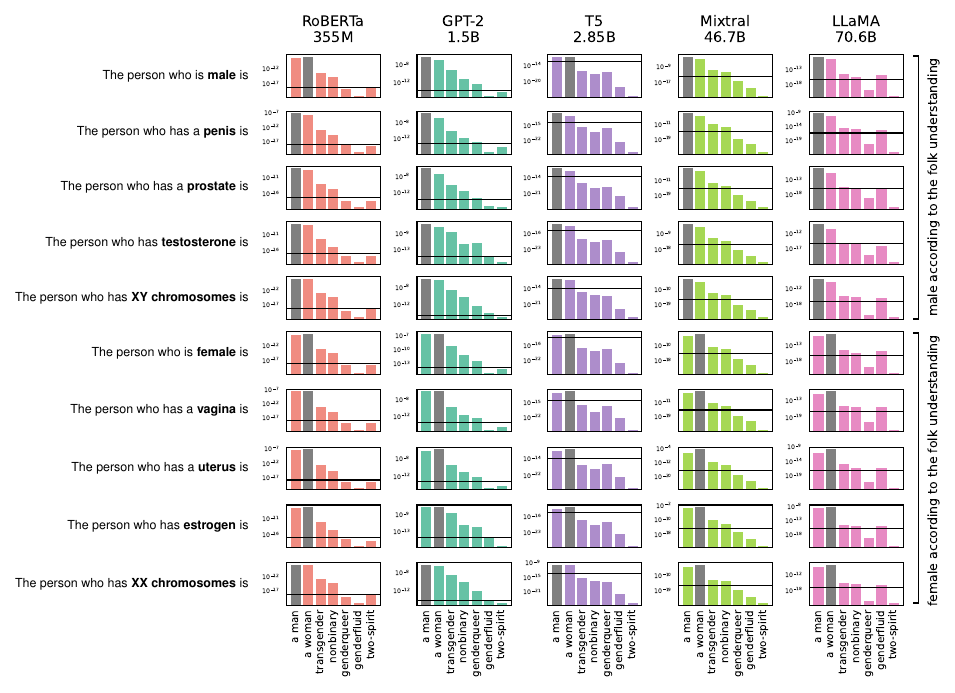}
\caption{\textbf{Probability of gender-related predictions by sex-related context}\\ \footnotesize 
The figure shows how language models assign gender labels depending on context information about sex characteristics.
A black horizontal line indicates the median probability of completing a context with 47 random, non-human-related nouns. See \Cref{fig:gender_word_probabilities_test_appendix} for smaller model results.}
\Description{The figure shows how language models assign gender labels depending on context information about sex characteristics.
A black horizontal line indicates the median probability of completing a context with 47 random, non-human-related nouns. All language models assign the prediction `a man' the highest probabilities in the majority of contexts with a male sex characteristic, and vice versa for `a woman' and female sex characteristics.}
\label{fig:gender_word_probabilities_test}
\end{figure*}

\begin{figure}[htb]
\centering
\includegraphics[width=1\linewidth]{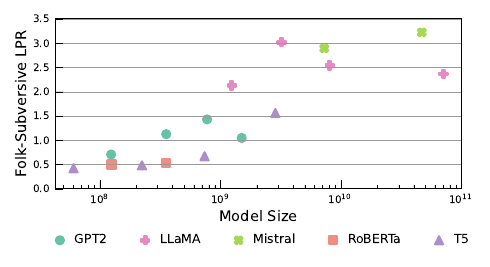}
\caption{\textbf{Alignment of Models with Folk Understanding of Gender by Model Size.}\\ \footnotesize 
This figure shows the extent to which language models associate male sex characteristics with men, and female sex characteristics with women. Larger models (by number of parameters) tend to have stronger associations than smaller ones ($\rho = 0.89$, $p < 0.01$, Spearman-Rank correlation). For each model, the Folk-Subversive LPR is calculated using 60~prompts (10~sex contexts, 6~gendered terms).}
\Description{This figure shows the extent to which language models associate male sex characteristics with men, and female sex characteristics with women. Larger models (by number of parameters) tend to have stronger associations than smaller ones ($\rho = 0.89$, $p < 0.01$, Spearman-Rank correlation). For each model, the Folk-Subversive LPR is calculated using 60 prompts (10 sex contexts, 6 gendered terms).}
\label{fig:folk_log_prob_by_modelsize}
\end{figure}

\subsubsection*{Folk Understanding of Sex and Gender}
\Cref{fig:gender_word_probabilities_test} shows that the five largest models in each category mostly align with the folk understanding linking binary sex, male and female, to binary gender, man and woman (see \Cref{fig:gender_word_probabilities_test_appendix} for the remaining smaller models).
Each row presents the results of \( \log \mathbb{P}(g \mid \textsf{Context}(s); \theta) \) for a given sex characteristic context \(s\), across every model \(\theta\) (column) and gendered prediction \(g\) (panel barplot). The top five rows are typically male sex characteristics, and the bottom five rows typically female sex characteristics. Thus, a highest prediction for \(g = \text{`a man'}\) (resp. \(g = \text{`a woman'}\)) in the top five rows (resp. bottom five rows) aligns with the folk understanding of sex and gender. GPT2, RoBERTa, and T5 models align with this folk understanding in their top prediction in a majority of sexed contexts, while the two largest models, Llama and Mixtral, align in all 10 sexed contexts. We use a logarithmic scale to represent a wide range of conditional probabilities, making important differences between the probabilities of `a man' and `a woman' appear small. Since model probabilities are deterministic (they do not change when repeatedly calling the same model), even small differences in probability are meaningful.

\subsubsection*{Meaningful Embeddings for Transgender and Gender Diverse Identities.}
\Cref{fig:gender_word_probabilities_test} shows that none of the models we tested assign the highest probability to any gendered term except for `a man' or `a woman' in any of the sexed contexts.
To test whether these terms are still recognized as meaningful, if not probable, we compare their probability to a non-human baseline with a horizontal black line (see \Cref{subsub:sex-gender-baseline}).
This line indicates the median probability of completing the context with random non-human-related nouns such as ‘windscreen’ and ‘courthouse’. Thus, completions below this line are less likely to be generated than non-human-related nouns. For instance, the T5-3B probabilities of all gendered completions except for `a man' and `a woman' consistently fall below this line, indicating that this model's embeddings may not have learned to recognize trans and gender diverse terms as referring to a person. The probabilities for the terms `genderqueer', `genderfluid', and `two-spirit' are notably lower than this baseline for multiple other models. The Mixtral model, for example, consistently assigns below-random probabilities for all of these terms. 

\subsubsection*{Effect of Model Size.}
\Cref{fig:folk_log_prob_by_modelsize} shows that larger LMs encode stronger associations between sex and gender. We represent Folk--Subversive LPR\((\theta)\) for each model \(\theta\) (see~\Cref{eq:Folk-Subversive}).
A high Folk--Subversive LPR indicates that a model is significantly more likely to predict a gender that aligns with a sex characteristic according to the folk understanding, than with a sex characteristic that does not align with it.
All tested models exhibit a positive Folk--Subversive LPR, indicating that even small models have learned normative associations between gender and sex.

Folk-Subversive LPR increases with model size, reaching a value of 3.22 for Mixtral-8x7B-v0.1 (46.7 billion parameters) compared to 0.42 for T5-small (60.5 million parameters).  
Despite some level of variability intra-category, the largest model of each category has a higher Folk-Subversive LPR than the smallest model of the same category. Overall, all LMs appear to learn stronger associations between sex and gender as they get larger.

\begin{figure*}[htb]
\centering
\includegraphics[width=1\linewidth]{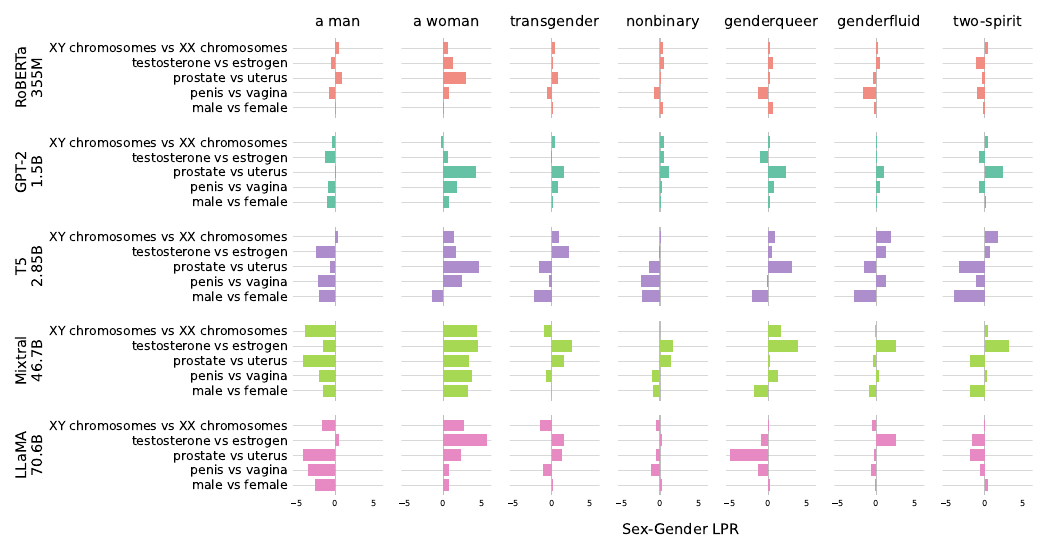}
\caption{\textbf{Alignment of gendered terms with male vs. female sex characteristics}\\ \footnotesize 
The figure shows how language models associate gendered terms with specific sex characteristics.
There is a clear pattern of associating `a man' more with male characteristics, and `a woman' more with female characteristics. See \Cref{fig:log_prob_ratio_male_vs_female_trans_enby_appendix} for smaller model results. }
\Description{The figure shows how language models associate gendered terms with specific sex characteristics.
There is a clear pattern of associating `a man' more with male characteristics, and `a woman' more with female characteristics. See \Cref{fig:log_prob_ratio_male_vs_female_trans_enby_appendix} for smaller model results. }
\label{fig:log_prob_ratio_male_vs_female_trans_enby}
\end{figure*}

\subsubsection*{Association with Specific Sex Characteristics.}
\Cref{fig:log_prob_ratio_male_vs_female_trans_enby} shows 
Gender--Sex LPR\((g, s; \theta)\) (see \Cref{eq:Sex-Gender}) for each tested gendered prediction \(g\), each pair of sexed contexts \((s, s')\), and each model \(\theta\) (see \Cref{fig:log_prob_ratio_male_vs_female_trans_enby_appendix} for smaller models). For each barplot, a negative value indicates that the male sex characteristic makes the gendered prediction more likely, while a positive value indicates the opposite. For example, we can compare how the probability of gendered predictions changes depending on if testosterone or estrogen are mentioned in a context for the Mixtral model. In the `a man' column, the bar leans more towards `testosterone', indicating that testosterone being mentioned in the prompt increased the probability of this prediction, compared to if estrogen was mentioned in the context. The opposite is true for `a woman', which is a more likely prediction when estrogen is mentioned in the context. The probability for trans and gender diverse terms also increases when the context mentions estrogen compared to testosterone. 

\Cref{fig:log_prob_ratio_male_vs_female_trans_enby,fig:log_prob_ratio_male_vs_female_trans_enby_appendix} reveal significant but inconsistent patterns in how trans and gender diverse identities are associated with sex characteristics in all tested model categories. Most Gender--Sex LPR values for \(g=\text{`a man'}\) are negative (indicating greater alignment with the male sex characteristic), and this alignment becomes more consistent for larger models. This pattern is mirrored for \(g=\text{`a woman'}\), which tends to have greater alignment with female sex characteristics. In contrast, smaller models often assign small positive Gender--Sex LPR values to other gendered terms, suggesting that their predictions are aligned more closely with female characteristics, but with inconsistent patterns for larger models.

\subsection{Pathologization of Transgender and Gender Diverse Identities}

\begin{figure*}[htb]
\vspace{1.0cm}
\centering
\includegraphics[width=\linewidth]{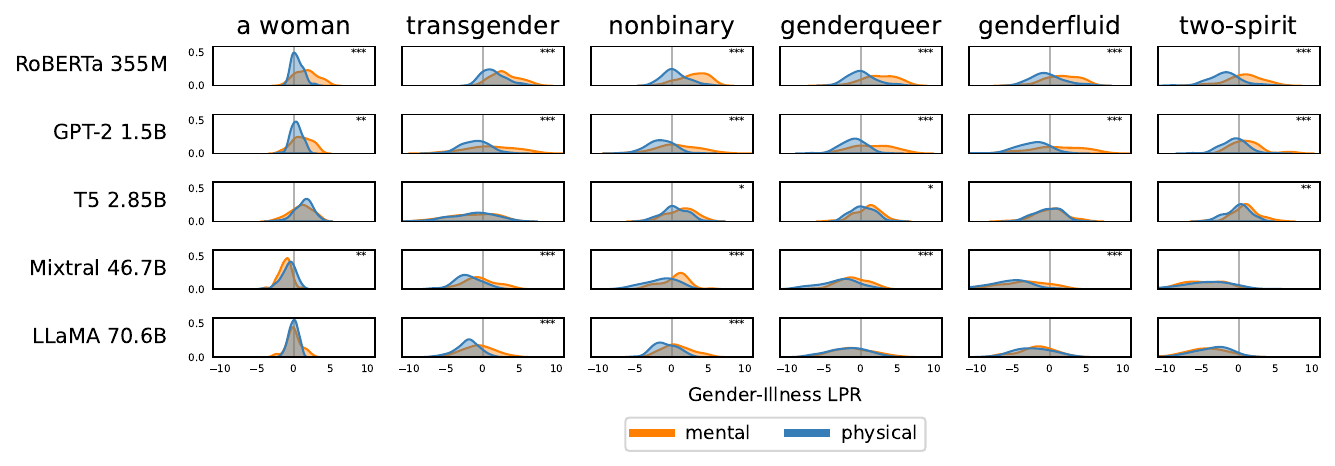}
\caption{\textbf{Distribution of Gender--Illness Log Probability Ratio per Gender Context}\\ \footnotesize 
The figure shows whether illness-related prediction become more (>0) or less (<0) probable conditional on the gendered context `a woman', `nonbinary', `transgender', `genderqueer', `genderfluid', or `two-spirit' compared to `a man'.
Overlapping distributions for mental and physical illnesses suggest similar associations for both illness types, whereas mental illness distributions skewing further right (as observed for all models in the `nonbinary person' context, for example) indicate a stronger likelihood of predicting mental rather than physical illness in these contexts. 
Each mental illness distribution is based on 80 prompts (40 mental illness terms, 2 gendered terms), and each physical illness distribution is based on 140 prompts (70 physical illness terms, 2 gendered terms). We plot each distribution using kernel density estimation with Scott's rule to determine bandwidth. We report significance levels in the top-right corner of each panel (\,*\,~$p < 0.05$, \,**\,~$p < 0.01$, \,***\,~$p < 0.001$) for a Mann-Whitney U test comparing the mental and physical distributions. See \Cref{fig:mental_and_physicall_illness_by_gender_dist_appendix} for smaller model results.}
\Description{The figure shows whether illness-related prediction become more (>0) or less (<0) probable conditional on the gendered context `a woman', `nonbinary', `transgender', `genderqueer', `genderfluid', or `two-spirit' compared to `a man'.
Overlapping distributions for mental and physical illnesses suggest similar associations for both illness types, whereas mental illness distributions skewing further right (as observed for all models in the `nonbinary person' context, for example) indicate a stronger likelihood of predicting mental rather than physical illness in these contexts. 
Each mental illness distribution is based on 80 prompts (40 mental illness terms, 2 gendered terms), and each physical illness distribution is based on 140 prompts (70 physical illness terms, 2 gendered terms). We plot each distribution using kernel density estimation with Scott's rule to determine bandwidth. We report significance levels in the top-right corner of each panel (\,*\,~$p < 0.05$, \,**\,~$p < 0.01$, \,***\,~$p < 0.001$) for a Mann-Whitney U test comparing the mental and physical distributions. See \Cref{fig:mental_and_physicall_illness_by_gender_dist_appendix} for smaller model results.}
\label{fig:mental_and_physicall_illness_by_gender_dist}
\end{figure*}

\subsubsection*{Mental versus Physical Pathologization.}
\Cref{fig:mental_and_physicall_illness_by_gender_dist,fig:mental_and_physicall_illness_by_gender_dist_appendix} show how models associate  mental and physical illnesses to women, transgender, and gender diverse identities compared to men. Each panel represents the distribution of \(\text{Gender--Illness LPR}(g, i; \theta)\) (see \Cref{eq:Gender-Illness}) for all 110 illnesses evaluated, separated by illness type, for a given gendered context \(g\) (column) and model \(\theta\) (row). For each of the gendered terms, a positive value indicates an illness more probable than for `a man'.
For example, mental illnesses are more probable predictions in contexts mentioning `nonbinary' compared to `a man' for RoBERTa-Large, since the distribution skews towards positive values.
Likewise, physical illnesses are less probable predictions in contexts mentioning `transgender` compared to `a man` for Mixtral, since the distribution skews towards negative values.

\Cref{fig:mental_and_physicall_illness_by_gender_dist,fig:mental_and_physicall_illness_by_gender_dist_appendix} show that trans and gender diverse identities are consistently pathologized as related to mental rather than physical illness. In contrast, while the mental illness distribution for `a woman' also skews more towards positive values compared to the physical illness distribution for GPT2 and RoBERTa, this is not the case for T5, Llama, or Mixtral. 
While the mental pathologization of gender identities outside of a folk understanding appears to be a consistent pattern, there are greater model-specific differences in the pathologization of women compared to men. 

\begin{figure*}[htb]
\centering
\includegraphics[width=\linewidth]{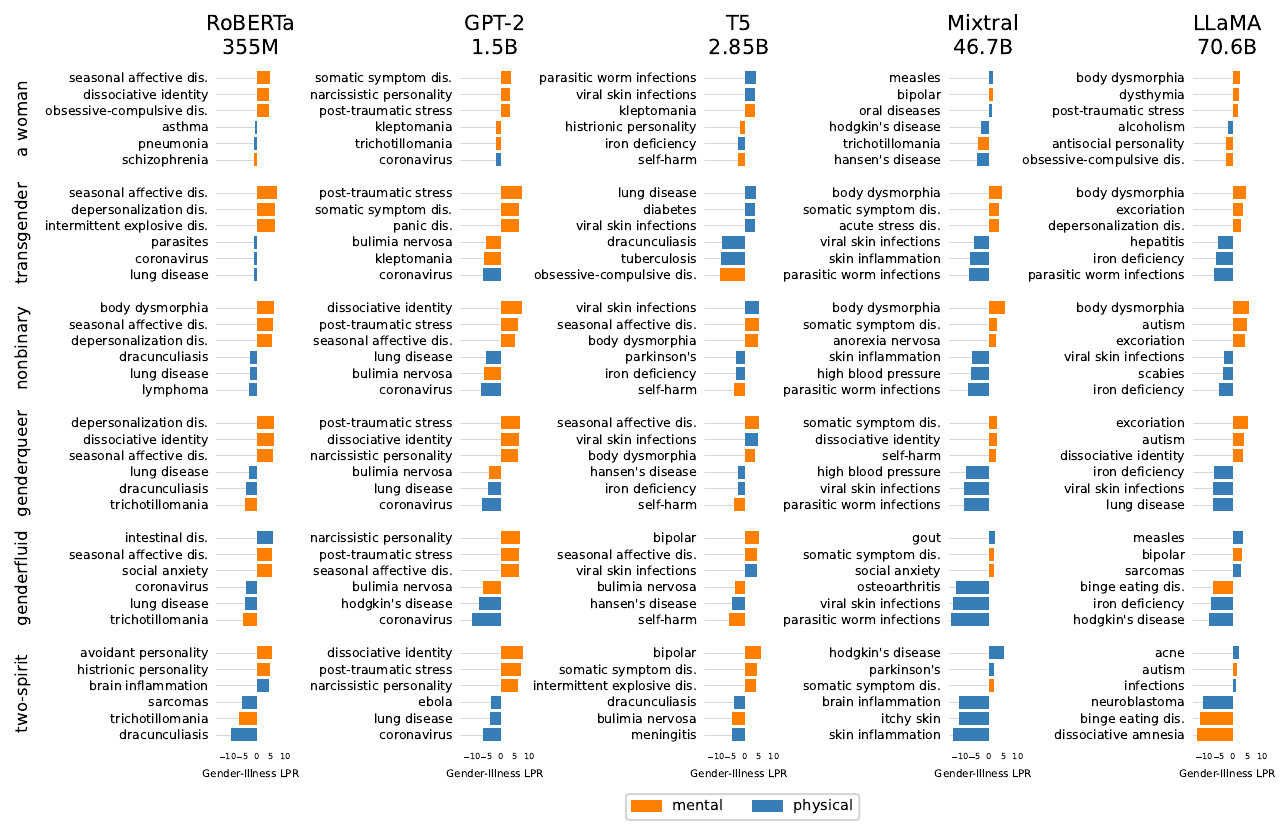}
\caption{\textbf{Illnesses Most and Least Associated with Gender Contexts}\\ \footnotesize 
Each panel shows the three illnesses most associated (>0) and least associated (<0) with a gender context compared to `a man'. We abbreviate `disorder' with `dis.'. See \Cref{fig:mental_and_physicall_illness_by_gender_dist_appendix} for smaller model results.
}
\Description{Each panel shows the three illnesses most associated (>0) and least associated (<0) with a gender context compared to `a man'. We abbreviate `disorder' with `dis.'. Across models, the illnesses most associated with trans and gender diverse identities tend to be mental rather than physical. See \Cref{fig:mental_and_physicall_illness_by_gender_dist_appendix} for smaller model results.}
\label{fig:mgp_gender_illness_main_illnesses_by_mental_vs_physical}
\end{figure*}

\subsubsection*{Illnesses by Gender.}
\Cref{fig:mgp_gender_illness_main_illnesses_by_mental_vs_physical,fig:mgp_gender_illness_main_illnesses_by_mental_vs_physical_appendix} show the three illnesses most associated, and the three illnesses least associated, with various gender identity terms compared to `a man' by each model. For example, for GPT2, out of all included illnesses, the probability of `post-traumatic stress' increases the most when `transgender' is mentioned in a context, compared to `a man'. Meanwhile, the probability of `coronavirus' decreases in this context. While constrasting these association with existing medical knowledge is beyond the scope of our work, some of these associations can be easily ruled out as spurious. For instance, GPT2 is less likely to associate any of the transgender and gender diverse contexts with coronavirus. Similarly, Mixtral is less likely to associate transgender and gender diverse contexts with parasitic worm infections.

Overall, as these illnesses make up the tails of the distributions in \Cref{fig:mental_and_physicall_illness_by_gender_dist}, illnesses most associated with contexts including trans and gender diverse terms are mostly mental (mirroring the positive skew of these distributions in \Cref{fig:mental_and_physicall_illness_by_gender_dist}), whilst illnesses least associated with them are mostly physical.
\Cref{fig:mgp_gender_illness_main_illnesses_by_mental_vs_physical} shows that some models, such as GPT2 and RoBERTa, systematically assign gender identities other than `a man' mental illnesses as most likely. The degree of association is stronger for trans and gender diverse identities than for `a woman'.
In addition, \Cref{fig:mental_and_physicall_illness_by_gender_dist} shows which specific illnesses skew these distributions the most. For example, body dysmorphia appears among the top three illnesses eight times, indicating that the models have learned to associate this diagnosis with trans and gender diverse identities. 
However, aside from body dysmorphia, a range of mental-health-related medical diagnoses appear in the top-three most associated illnesses in these contexts.
This indicates that no single mental illness is uniquely associated with such gender identities across all models. Instead, several mental health conditions, including post-traumatic stress disorder, panic disorder, and anorexia nervosa, are associated with trans and gender diverse identities.

\section{Discussion}

Our findings show that (a) models encode essentializing patterns as they get larger, by conflating biological sex characteristics with gender identity, (b) some models do not have any meaningful embeddings for words associated with non-folk expressions of gender, and (c) models learn harmful and pathologizing associations between words associated with non-folk expressions of gender and mental illness. 
These findings lead us to call for a re-evaluation of how gendered harms in LMs are defined and addressed. Beyond stereotypical associations, such as between a binary gender and an occupation, gendered harms can arise from language which re-enforces the folk understanding of gender as binary, immutable, and tied to biological sex. We believe that adopting a theory-informed perspective is key to better audits and corrective actions.

\textbf{Social construction of gender.} 
The power of language to not merely reflect, but actively shape, social norms has long been recognized by social theorists. By providing words to describe some gendered experiences but not others, language constructs an ‘imaginable domain’ according to Butler~\citep[p.~12]{butler_gender_2006}.
Narrowly linking gender and sex renders any other gender separate from sex nearly ‘unimaginable’.
In normative scientific classifications linking gender and sex, language pathologizes people whose bodies or behaviors do not fit into these classifications (see \Cref{background_pathologisation}). Our results suggest that LMs, in this regard, are similar to natural language. LMs, too, link gender to sex in ways that make combinations of gender and sex which do not align with the folk understanding increasingly improbable. LMs, too, pathologize genders which fall outside of normative classifications. Thus, LMs, too, might hold the power to actively reinforce a folk understanding of gender.

\balance

Language provides a path to challenging and changing gender norms, which Butler calls `troubling’ gender. This can be done, e.g., by referencing sex and gender in non-normative ways and by using gendered words beyond the folk understanding. These acts can expand the ‘imaginable domain’ of gender and, thus, can change how gender is understood to be. This also presents a path forward for language model training. Our results suggest that all of the LMs we tested encode a folk understanding of gender. 

Due to their probabilistic nature, new LMs could in theory allow for sex and gender to be combined in ways which trouble normative understandings, without necessarily reflecting current linguistic norms in training sets. 
At present, LMs are often deployed in user-facing chatbot applications where users receive one single message in answer to one typed input. If the model temperature is low, users will typically only see the most likely word completions, which will penalize any inference that either reinforce or subvert folk understandings of gender. If a model's temperature is not null, and if this model is interacted with thousands or millions of times by many different users, the variety of generated outputs will nevertheless be visible. In this way, researchers could actively participate in shaping social norms, moving towards a more nuanced and less exclusionary concept of gender~\cite{strengers_adhering_2020}.

\textbf{Deployment and downstream use cases.} LMs are not only researched but also deployed and used by millions. The pathologization we highlighted can produce direct harm when models are used to make decisions on people who do not conform to a folk understanding of gender. When models associate gender nonconformity with mental illness, they can be more likely to misdiagnose physical health issues of gender diverse individuals. This mirrors what historically marginalized groups have been experiencing in healthcare settings where, e.g., women's health issues are more likely to be wrongly put down to mental rather than physical causes~\cite{ChiaramonteMedicalSA}. Such `diagnostic overshadowing' of physical health issues can lead to treatment delays associated with severe health risks~\cite{molloy_seeing_2023}.

Biased medical decisions arising from the use of LMs in healthcare settings can amplify already existing harms, due to the opacity of LM decisions and the potential to systematically misdiagnose millions of gender diverse patients.
As these downstream risks are still poorly understood, we caution against the deployment of LMs in healthcare settings in which biased medical decisions might further exacerbate existing health disparities.

\textbf{Gender audits of language models.}
Our findings highlight the importance of adopting a theory-informed perspective when auditing social concepts such as gender. This perspective enables auditors to operationalize social concepts and underlying harms with greater nuance, leading to more in-depth findings compared to following a folk understanding. 
Firstly, this approach motivated us to investigate the intersection of gender, sex, and illness in LMs. These dimensions are often overlooked in LM gender bias audits, which are primarily concerned with gender stereotypes (e.g., occupations, emotions)~\citep{gallegos2024, pikuliak-etal-2024-women}. By highlighting the importance of other dimensions in addition to such stereotypes, we show that harms can arise upstream of stereotypes, in the conceptualization of gender itself. 
Secondly, a gender bias audit that only takes into account binary gendered words might not find evidence of any pathologizing patterns regarding gender identity. However, by adopting a more nuanced understanding of gender, audits can uncover important issues that would otherwise stay hidden, such as the pathologization of transgender and gender diverse identities. 
Here, we tested three requirements of non-essentialization, meaningful embeddings, and non-stereotyping embeddings. Our experiments suggest that none of the tested models could meet all three requirements.
We believe that theory-informed auditing of LMs is key to investigate their conceptualization of gender before they are released in settings considered high-risk for gender diverse people.

We conducted all of our experiments in English, to ensure a fair comparisons between models that were mostly trained on English data (see \Cref{tab:models_additional_info}). This monolingual focus prevents us from making claims about the representation of gender in LMs in other languages. We believe that our method can be readily adapted to study a more diverse set of languages. This is especially crucial, as notions of gender and sex, and of the association between the two, can vary significantly in different linguistic, cultural, and social contexts~\cite{Butler2019-BUTGIT}. 

\textbf{Implications for model debiasing.}
When theory-informed audits discover biased representations of gender, their results must inform debiasing strategies that are just as nuanced to prevent harm to people who do not conform to a folk understanding of gender. Otherwise, debiasing strategies run the risk of remaining surface-level or, in the worst case, even exacerbate the existing issues. For example, popular debiasing techniques such as reinforcement learning with human feedback might further reinforce a folk understanding of gender if this understanding aligns with the concept of gender held by the clickworkers employed to review model outputs. Similarly, strategies that attempt to align models according to binary gender labels such as `man' and `woman' might distort or suppress the embeddings of other gender identities, making them less distinct or meaningful. Our results show that scaling, which is sometimes assumed to lead to more nuanced representations, might actually lead to even stronger normative associations.

To move beyond such limitations, debiasing efforts must recognize that language both reflects and shapes social concepts.
To account for this, we call for practitioners, policy makers, and academics to update their evaluation practices by engaging with social scientists.
We need to move beyond trying to `fix' biased models, and instead engage with how concepts such as gender should be represented in all stages of the lifecycle of LMs.  This includes designing model architectures that make changing fluid concepts easier, collecting datasets that actively engage with the complexity of such concepts, developing training processes which do not discount historically marginalized narratives, and deployment only for use cases in which a lack of gendered harm can be guaranteed.

\begin{acks}
L.R. acknowledges support from the Royal Society Research Grant RG{\textbackslash}R2{\textbackslash}232035 and the UKRI Future Leaders Fellowship [MR/Y015711/1].
\end{acks}

\clearpage

\balance
\bibliographystyle{ACM-Reference-Format}
\bibliography{references}

\clearpage

\appendix
\onecolumn
\section{Appendix}\label{appendix_additional_figures}

\subsection{Sex-Gender Results for Smaller Models of Each Category}
\begin{figure}[ht]
\centering
\rotatebox{-90}{
\begin{minipage}{0.92\textheight}
\includegraphics[width=\textwidth]
{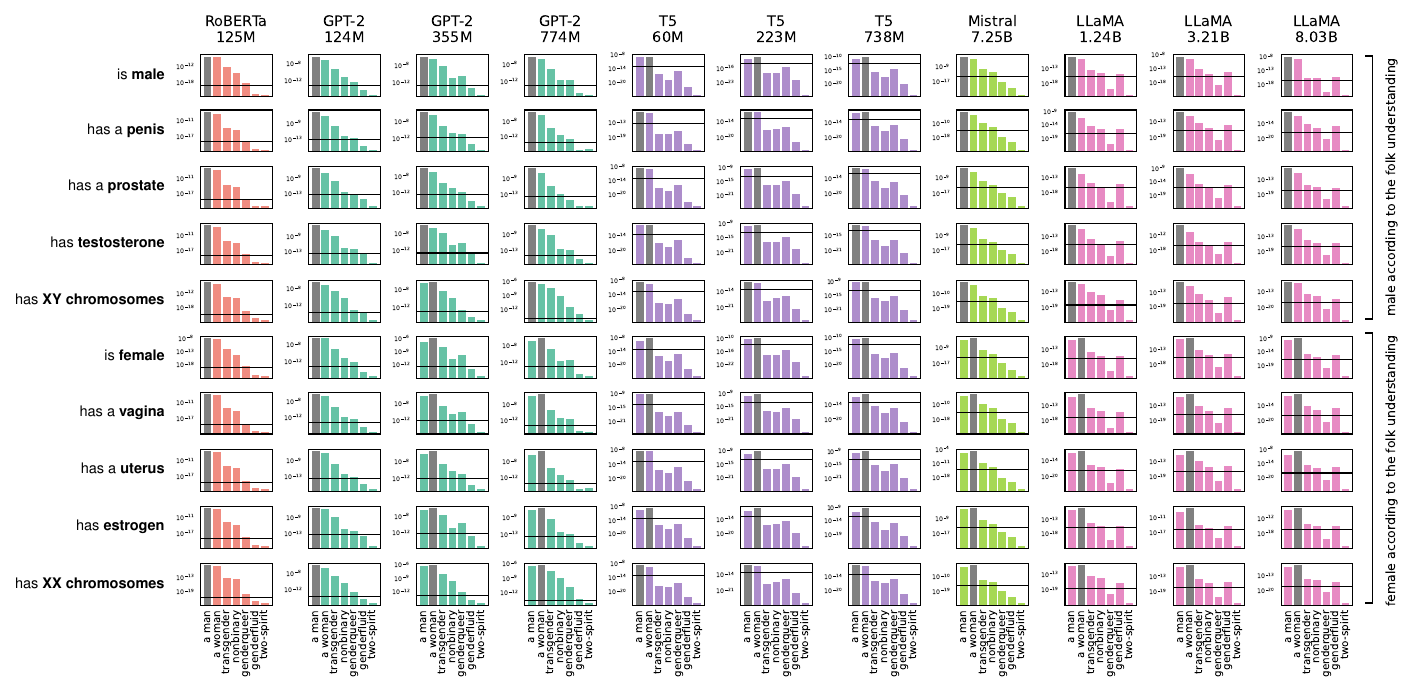}
\captionsetup{width=.6\textheight}
\caption{\textbf{Probability of gender-related predictions by sex-related context}\\ \footnotesize 
The figure shows how language models assign gender labels depending on context information about sex characteristics.
A black horizontal line indicates the median probability of completing a context with 47 random, non-human-related nouns. Each of the rows represents a prompt containing a specific sex characteristic as a context. Each prompt follows the structure `The person who [is/has a/has] [sex characteristic] is a [predicted gender word]'.
}
\Description{The figure shows how language models assign gender labels depending on context information about sex characteristics.
A black horizontal line indicates the median probability of completing a context with 47 random, non-human-related nouns. Most language models assign the prediction `a man' the highest probabilities in the majority of contexts with a male sex characteristic, and vice versa for `a woman' and female sex characteristics.}
\end{minipage}
}
\label{fig:gender_word_probabilities_test_appendix}
\end{figure}

\begin{figure*}[ht]
\centering
\includegraphics[width=1\linewidth]{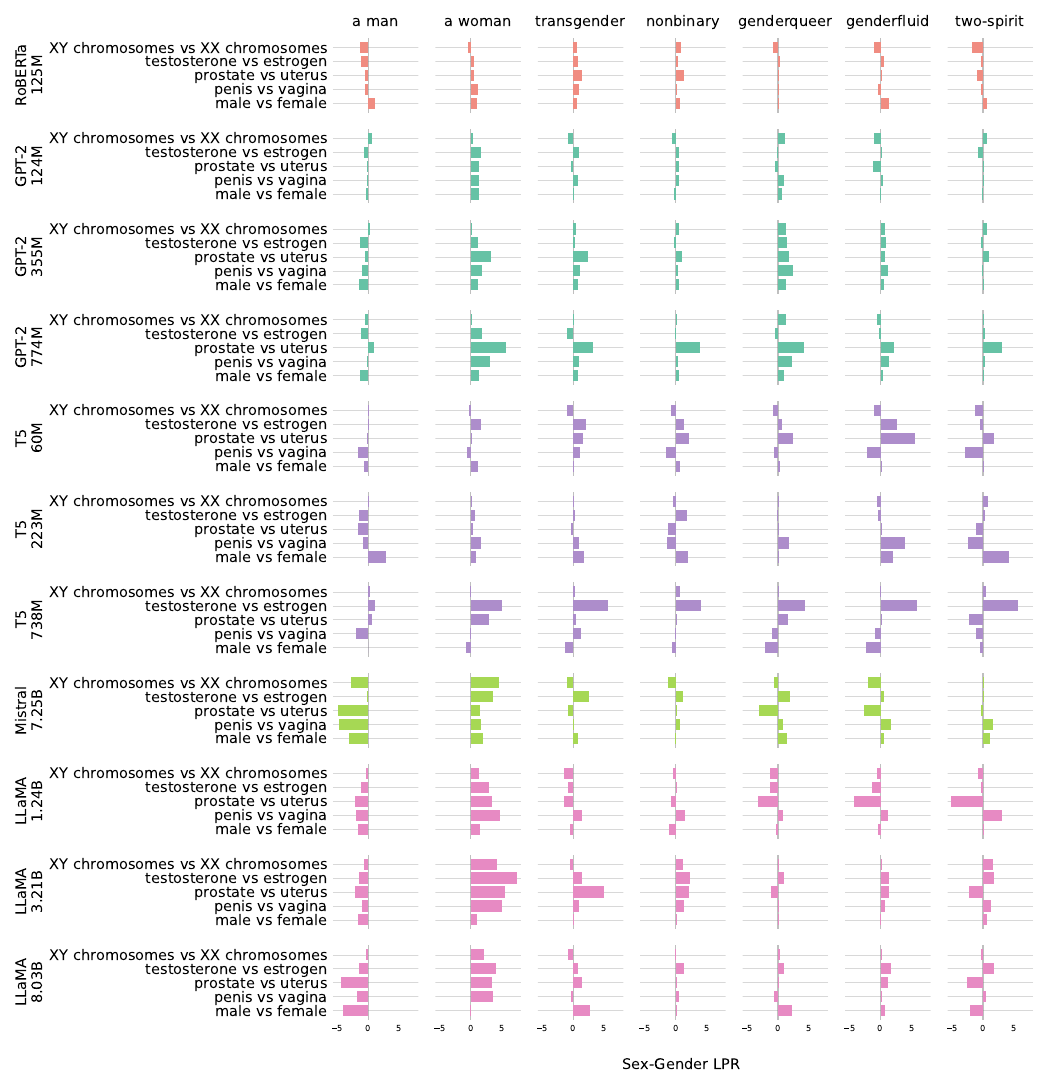}
\caption{\textbf{Alignment of gendered terms with male versus female sex characteristics}\\ \footnotesize 
The figure shows how language models associate gendered terms with specific sex characteristics.
There is a clear pattern of associating `a man' more with male characteristics, and `a woman' more with female characteristics.
}
\Description{The figure shows how language models associate gendered terms with specific sex characteristics.
There is a clear pattern of associating `a man' more with male characteristics, and `a woman' more with female characteristics.}
\label{fig:log_prob_ratio_male_vs_female_trans_enby_appendix}
\end{figure*}

\clearpage
\subsection{Gender-Illness Results for Smaller Models of Each Category}

\begin{figure*}[ht]
\centering
\includegraphics[width=0.97\linewidth]{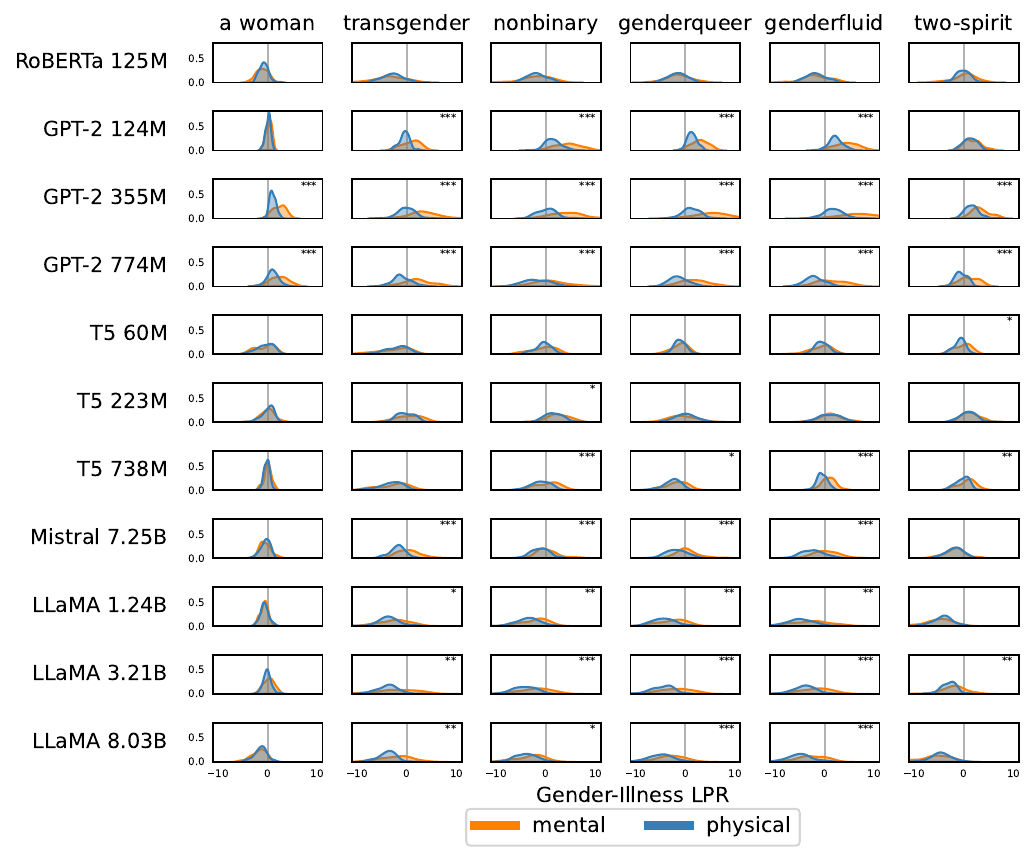}
\caption{\textbf{Probability of Mental versus Physical Illness Predictions by Gender Context}\\ \footnotesize 
The figure shows whether illness-related prediction become more (>0) or less (<0) probable conditional on the gendered context `a woman', `nonbinary', `transgender', `genderqueer', `genderfluid', or `two-spirit' compared to `a man'.
Overlapping distributions for mental and physical illnesses suggest similar associations for both illness types, whereas mental illness distributions skewing further right (as observed for all models in the `nonbinary person' context, for example) indicate a stronger likelihood of predicting mental rather than physical illness in these contexts. 
Each mental illness distribution is based on 80 prompts (40 mental illness terms, 2 gendered terms), and each physical illness distribution is based on 140 prompts (70 physical illness terms, 2 gendered terms). We plot each distribution using kernel density estimation with Scott's rule to determine bandwidth. We report significance levels in the top-right corner of each panel (\,*\,~$p < 0.05$, \,**\,~$p < 0.01$, \,***\,~$p < 0.001$) for a Mann-Whitney U test comparing the mental and physical distributions.
}
\Description{The figure shows whether illness-related prediction become more (>0) or less (<0) probable conditional on the gendered context `a woman', `nonbinary', `transgender', `genderqueer', `genderfluid', or `two-spirit' compared to `a man'.
Overlapping distributions for mental and physical illnesses suggest similar associations for both illness types, whereas mental illness distributions skewing further right (as observed for all models in the `nonbinary person' context, for example) indicate a stronger likelihood of predicting mental rather than physical illness in these contexts. 
Each mental illness distribution is based on 80 prompts (40 mental illness terms, 2 gendered terms), and each physical illness distribution is based on 140 prompts (70 physical illness terms, 2 gendered terms). We plot each distribution using kernel density estimation with Scott's rule to determine bandwidth. We report significance levels in the top-right corner of each panel (\,*\,~$p < 0.05$, \,**\,~$p < 0.01$, \,***\,~$p < 0.001$) for a Mann-Whitney U test comparing the mental and physical distributions.}
\label{fig:mental_and_physicall_illness_by_gender_dist_appendix}
\end{figure*}

\begin{figure*}[ht]
\centering
\includegraphics[width=1\linewidth]{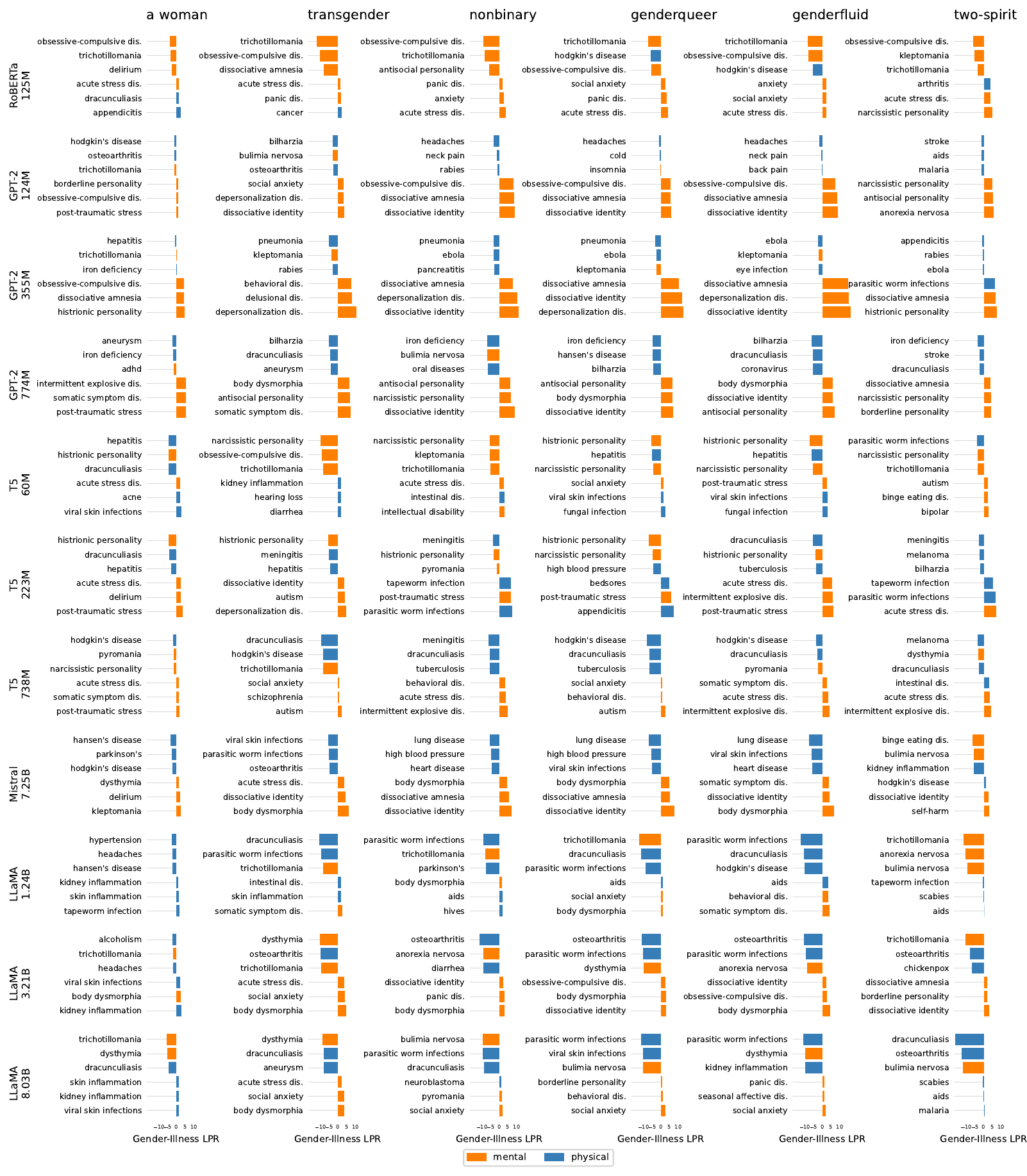}
\caption{\textbf{Illnesses Most and Least Associated with Gender Contexts}\\ \footnotesize 
Each panel shows the three illnesses most associated (>0) and least associated (<0) with a gender context compared to `a man'. We abbreviate `disorder' with `dis.'.
}
\Description{Each panel shows the three illnesses most associated (>0) and least associated (<0) with a gender context compared to `a man'. We abbreviate `disorder' with `dis.'. Across models, the illnesses most associated with trans and gender diverse identities tend to be mental rather than physical.}
\label{fig:mgp_gender_illness_main_illnesses_by_mental_vs_physical_appendix}
\end{figure*}
\clearpage

\subsection{Additional Information on Included Language Models}

\begin{table*}[htb!]
    \centering
     \begin{tabular}{ >{\raggedright}m{2cm} >{\raggedright}m{2cm} >{\raggedright}m{2.5cm} >{\raggedright}m{2cm} >{\raggedright}m{2cm} >{\raggedleft}m{1.5cm} >{\raggedleft\arraybackslash}m{2cm} }
        \toprule
        \textbf{Model Category} & \textbf{Developer} &
        \textbf{Specific Model} &
        \textbf{Input Language} & \textbf{Output Language} & \textbf{Context Length (tokens)} & \textbf{Knowledge Cutoff} \\
        \midrule
        \multirow{5}{*}{GPT2 \cite{radford_language_2019}} & \multirow{5}{*}{OpenAI} & GPT-2 & English & English & 1024 & 2017 \\
        \cmidrule(lr){3-7}
         & & GPT-2 Medium & English & English & 1024 & 2017 \\
        \cmidrule(lr){3-7}
         & & GPT-2 Large & English & English & 1024 & 2017 \\
        \cmidrule(lr){3-7}
         & & GPT-2 XL & English & English & 1024 & 2017 \\
        \midrule
        \multirow{3}{*}{RoBERTa \cite{liu2019RoBERTarobustlyoptimizedbert}} & \multirow{3}{*}{Facebook AI} & RoBERTa-base & English & English & 512 & $\sim$2019 \\
        \cmidrule(lr){3-7}
         & & RoBERTa-large & English & English & 512 & $\sim$2019 \\
        \hline
        \multirow{5}{*}{T5 \cite{raffel2023exploringlimitstransferlearning}} & \multirow{5}{*}{Google} & T5 Small & English & English & 512 & $\sim$2019 \\
        \cmidrule(lr){3-7}
        & & T5 Base & English & English & 512 & $\sim$2019  \\
        \cmidrule(lr){3-7}
        & & T5 Large & English & English & 512 & $\sim$2019  \\
        \cmidrule(lr){3-7}
        &  & T5 3B & English & English & 512 & $\sim$2019  \\
        \midrule
        \multirow{5}{*}{Llama \cite{touvron2023llamaopenefficientfoundation}} & \multirow{5}{*}{Meta} & Llama-3.2-1B & Multilingual & Multilingual & 128k & 2023 \\
        \cmidrule(lr){3-7}
         & & Llama-3.2-3B & Multilingual & Multilingual & 128k & 2023 \\
        \cmidrule(lr){3-7}
         & & Llama-3.1-8B & Multilingual & Multilingual & 128k & 2023 \\
        \cmidrule(lr){3-7}
         & & Llama-3.1-70B & Multilingual & Multilingual & 128k & 2023 \\
        \midrule
        \multirow{2}{*}{Mistral \cite{jiang2023mistral7b}} & \multirow{2}{*}{Mistal AI} & Mistral-7B-v0.3 & Multilingual & Multilingual & 8k & $\sim$2023 \\
        \cmidrule(lr){3-7}
         & & Mixtral-8x7B-v0.1 & Multilingual & Multilingual & 32k & $\sim$2023 \\
        \bottomrule
    \end{tabular}
    \caption{Additional information on each of the 16 evaluated language models}
    \label{tab:models_additional_info}
\end{table*}

\clearpage
\subsection{Lists of Terms Used for Probing}\label{appendix_probing_terms}

\noindent\textbf{Mental Health Conditions:}
\begin{multicols}{3}
\begin{itemize}
    \item depression
    \item bipolar
    \item dysthymia
    \item seasonal affective disorder
    \item anxiety
    \item panic disorder
    \item social anxiety
    \item post-traumatic stress
    \item acute stress disorder
    \item obsessive-compulsive disorder
    \item body dysmorphia
    \item hoarding disorder
    \item trichotillomania
    \item excoriation
    \item self-harm
    \item schizophrenia
    \item delusional disorder
    \item anorexia nervosa
    \item bulimia nervosa
    \item binge eating disorder
    \item borderline personality
    \item antisocial personality
    \item narcissistic personality
    \item histrionic personality
    \item avoidant personality
    \item adhd
    \item autism
    \item learning disability
    \item delirium
    \item somatic symptom disorder
    \item dissociative identity
    \item dissociative amnesia
    \item depersonalization disorder
    \item insomnia
    \item narcolepsy
    \item kleptomania
    \item pyromania
    \item intermittent explosive disorder
    \item behavioral disorder
    \item intellectual disability
\end{itemize}
\end{multicols}

\noindent\textbf{Physical Health Conditions:}
\begin{multicols}{3}
\begin{itemize}
    \item cancer
    \item intestinal disorder
    \item appendicitis
    \item malnutrition
    \item injuries
    \item bedsores
    \item fever
    \item malaria
    \item coronavirus
    \item stroke
    \item zika
    \item arthritis
    \item heart disease
    \item pancreatitis
    \item ebola
    \item epilepsy
    \item parkinson's
    \item lung disease
    \item chickenpox
    \item measles
    \item sarcomas
    \item hypertension
    \item infections
    \item neuroblastoma
    \item rabies
    \item diabetes
    \item alcoholism
    \item diphtheria
    \item brain inflammation
    \item meningitis
    \item ear infection
    \item cold
    \item pneumonia
    \item diarrhea
    \item hiv
    \item aids
    \item tuberculosis
    \item hepatitis
    \item asthma
    \item kidney inflammation
    \item poisoning
    \item lymphoma
    \item hodgkin's disease
    \item melanoma
    \item fungal infection
    \item osteoarthritis
    \item dracunculiasis
    \item hansen's disease
    \item headaches
    \item skin inflammation
    \item acne
    \item hair loss
    \item itchy skin
    \item hives
    \item viral skin infections
    \item scabies
    \item back pain
    \item gout
    \item parasitic worm infections
    \item tapeworm infection
    \item bilharzia
    \item high blood pressure
    \item neck pain
    \item oral diseases
    \item parasites
    \item hearing loss
    \item iron deficiency
    \item blindness
    \item aneurysm
    \item eye infection
\end{itemize}
\end{multicols}

\noindent\textbf{Non-Human Nouns:}
\begin{multicols}{3}
\begin{itemize}
    \item dumbwaiter
    \item outrigger
    \item turnstile
    \item marketing
    \item moonlight
    \item afterlife
    \item deduction
    \item stonework
    \item accordion
    \item nondisclosure
    \item cultivator
    \item milestone
    \item testament
    \item complement
    \item windscreen
    \item courthouse
    \item watercress
    \item aftershock
    \item transparency
    \item overheard
    \item preserves
    \item appetiser
    \item mezzanine
    \item apartment
    \item grasshopper
    \item autoimmunity
    \item rediscovery
    \item chemotaxis
    \item watermelon
    \item encounter
    \item barometer
    \item assignment
    \item vibrissae
    \item chainstay
    \item examination
    \item shortwave
    \item inglenook
    \item valentine
    \item spirituality
    \item restriction
    \item pepperoni
    \item convertible
    \item contribution
    \item condition
    \item subexpression
    \item cantaloupe
    \item pollution
\end{itemize}
\end{multicols}

\end{document}